%% file: habigt2013.tex
\documentclass{article}
\usepackage{spconf,amsmath,graphicx}
\usepackage[utf8]{inputenc}
\usepackage{color}
\usepackage[font=footnotesize]{caption}
\usepackage{subcaption}
\usepackage{tabularx}
\usepackage{pgfplots}
\usepackage[pdftitle={Image Completion for View Synthesis Using Markov Random Fields and Efficient Belief Propagation},pdfauthor={Julian Habigt, Klaus Diepold},pdfkeywords={DIBR, View Synthesis, Inpainting, Hole-Filling, MRF}]{hyperref}

\newlength\figureheight
\newlength\figurewidth

\title{\MakeUppercase{Image Completion for View Synthesis Using~Markov~Random~Fields~and~Efficient~Belief~Propagation}}
\name{Julian Habigt and Klaus Diepold}
\address{Technische Universität München, Institute for Data Processing,\\ Arcisstr. 21, 80333 Munich, Germany\\jh@tum.de, kldi@tum.de}
\begin{document}
\onecolumn
\ninept
\begin{center}
\begin{tabular}{|l|l|}
  \hline
  \multicolumn{2}{|c|}{\textbf{Image Completion for View Synthesis Using Markov Random Fields and Efficient Belief Propagation}} \\
  \multicolumn{2}{|c|}{Julian Habigt and Klaus Diepold} \\
  \hline
  \multicolumn{2}{|c|}{\begin{minipage}{5in}View synthesis is a process for generating novel views from a scene which has been recorded with a 3-D camera setup. It has important applications in 3-D post-production and 2-D to 3-D conversion. However, a central problem in the generation of novel views lies in the handling of disocclusions. Background content, which was occluded in the original view, may become unveiled in the synthesized view. This leads to missing information in the generated view which has to be filled in a visually plausible manner. We present an inpainting algorithm for disocclusion filling in synthesized views based on Markov random fields and efficient belief propagation. We compare the result to two state-of-the-art algorithms and demonstrate a significant improvement in image quality. \end{minipage}} \\
  \hline
  Published in: & Proc. IEEE International Conference on Image Processing (ICIP 2013) \\
  \hline
  Date of Conference: & 15-18 Sep. 2013 \\
  \hline
  Pages: & 2131 - 2134 \\
  \hline
  Publisher: & IEEE \\
  \hline
  DOI: & 10.1109/ICIP.2013.6738439 \\
  \hline
  WWW: & http://ieeexplore.ieee.org/xpl/articleDetails.jsp?arnumber=6738439 \\
  \hline
  IEEE Copyright Notice: & \begin{minipage}{5in}© 2013 IEEE. Personal use of this material is permitted. Permission from IEEE must be obtained for all other uses, in any current or future media, including reprinting/republishing this material for advertising or promotional purposes, creating new collective works, for resale or redistribution to servers or lists, or reuse of any copyrighted component of this work in other works.\end{minipage} \\
  \hline
\end{tabular}
\end{center}
\twocolumn
\maketitle
\begin{abstract}
View synthesis is a process for generating novel views from a scene which has been recorded with a 3-D camera setup. It has important applications in 3-D post-production and 2-D to 3-D conversion. However, a central problem in the generation of novel views lies in the handling of disocclusions. Background content, which was occluded in the original view, may become unveiled in the synthesized view. This leads to missing information in the generated view which has to be filled in a visually plausible manner. We present an inpainting algorithm for disocclusion filling in synthesized views based on Markov random fields and efficient belief propagation. We compare the result to two state-of-the-art algorithms and demonstrate a significant improvement in image quality.
\end{abstract}
\section{Introduction}
\label{sec:intro}
View synthesis is an important tool for the generation of content for 3-D television~\cite{Smolic2011}. In traditional stereoscopic setups, the scene is filmed with two cameras and then reproduced on a 3-D television screen, which can produce two separate pictures for the eyes of the viewer. This approach has several drawbacks. The baseline of the setup, i.e., the distance between the two cameras, has to be fixed during the production of the 3-D content and cannot be changed afterwards. When this content is shown on screens of different sizes, e.g., in a cinema or on a mobile device, the common baseline leads to an incorrect reproduction of the perceived depth of the scene~\cite{Held2008}. Furthermore, current autostereoscopic displays, i.e., displays which don't require the viewer to wear glasses to see 3-D content need a much higher number of views of the same scene, e.g., 28 or more. It is therefore necessary to be able to generate virtual views of a scene once the scene has been recorded.

One technique to generate such virtual views which has gained momentum in recent years is called depth image-based rendering. There, the virtual view is generated from the image of one or more cameras and corresponding depth maps. A central problem in the generation of novel views lies in the handling of disocclusions. Background content, which was occluded in the original view by objects that were closer to the camera, may become unveiled in the virtual view. In a setup with two or more cameras, these so-called disocclusions may be partially filled with content from another camera, yet some disocclusions usually still remain~\cite{Chen1993}. Even more challenging, when there is only one view and a corresponding depth map, there is no other other information available to fill the holes in the resulting view and the holes may have to be filled with synthetically generated content. This is, for example, the case in 2-D to 3-D conversion or in proposed transmission schemes with only one texture and depth map such as ATTEST~\cite{Fehn2004}.
\begin{figure}[tb]
\begin{subfigure}[b]{0.3\linewidth}
  \centering
  \includegraphics[width=\textwidth]{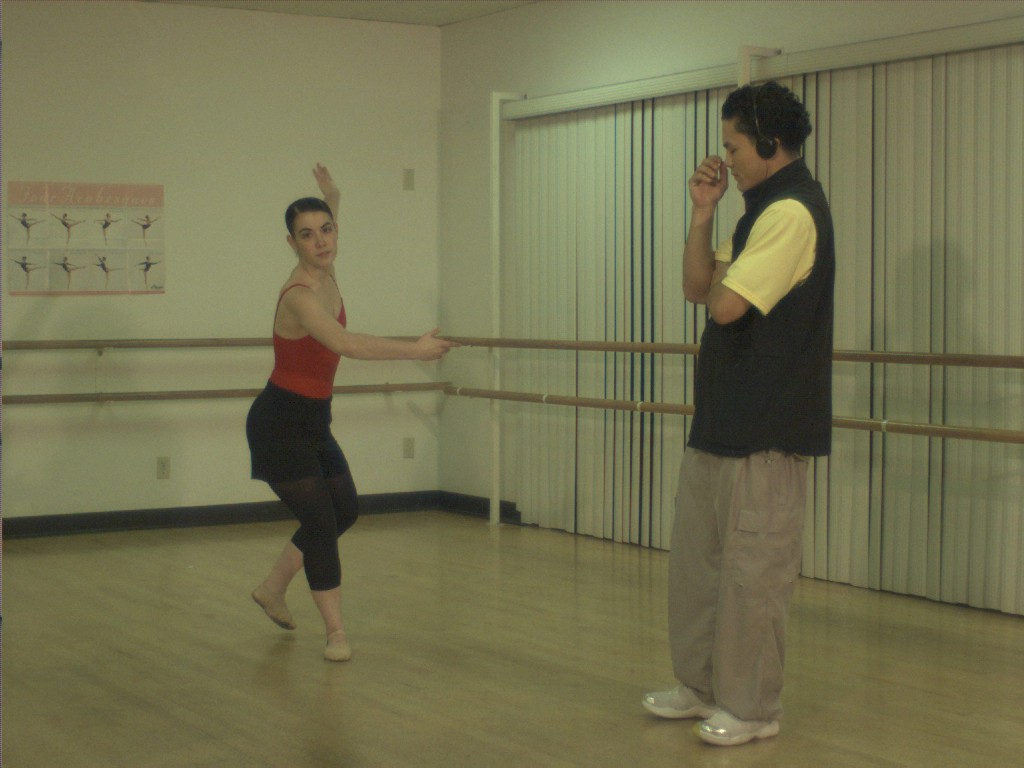}
\end{subfigure}\hfill
\begin{subfigure}[b]{.3\linewidth}
  \centering
  \includegraphics[width=\textwidth]{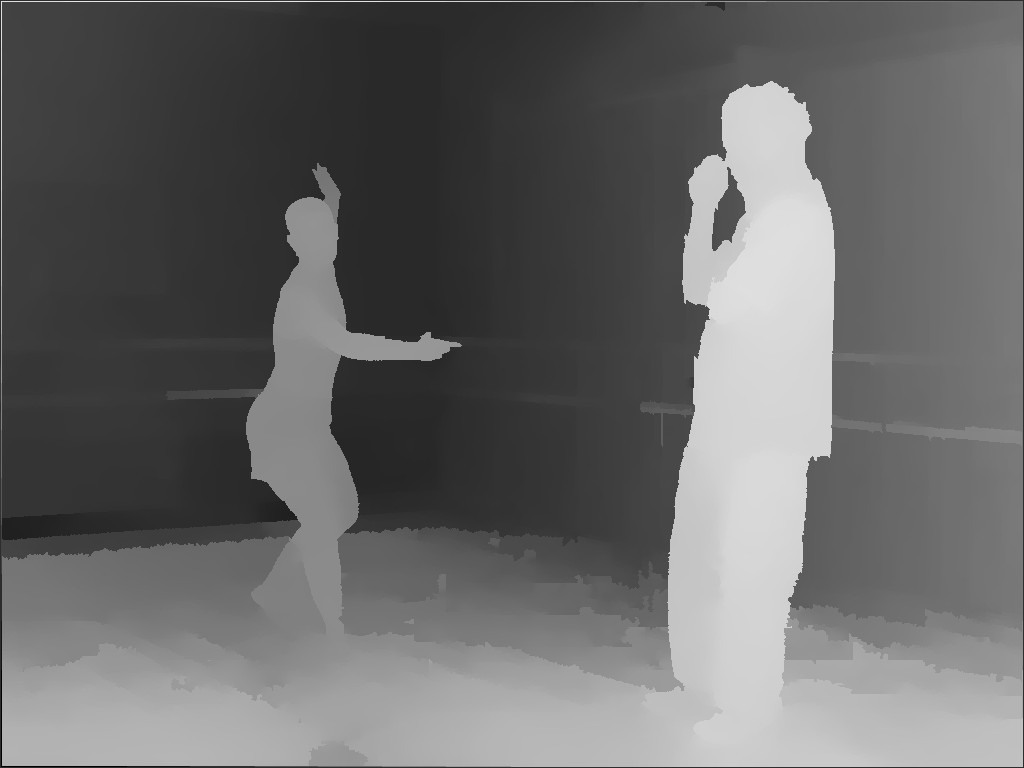}
\end{subfigure}\hfill
\begin{subfigure}[b]{.3\linewidth}
  \centering
  \includegraphics[width=\textwidth]{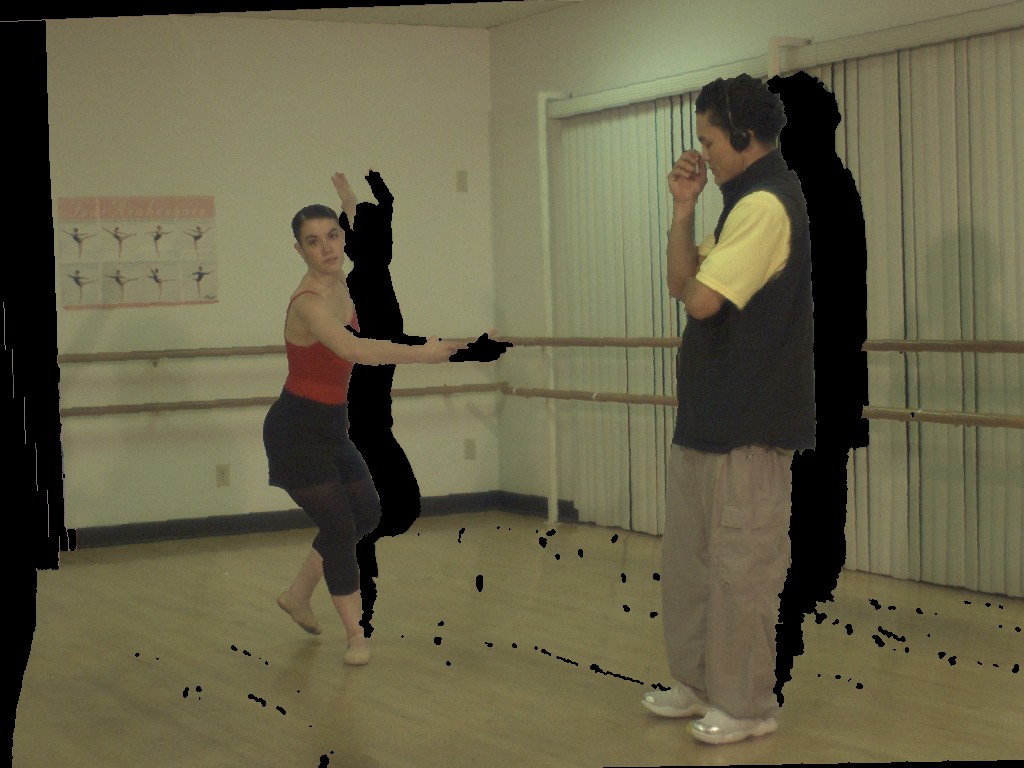}
\end{subfigure}
\caption{Given an input image and a corresponding depth-map, we can generate virtual views, a process which is called depth image-based rendering (DIBR). However, disocclusions appear where background is unveiled which was occluded by foreground objects in the original view.}
\label{fig:warping}
\end{figure}

A number of solutions have been proposed for this problem. One way to avoid it entirely would be the modification of the depth-map. Disocclusions appear at regions in the image where there is a steep gradient in the depth-map, i.e., at the borders of foreground objects. Zhang et al. \cite{Zhang2005} proposed a technique where the depth-map is filtered to remove these steep gradients. The result is a virtual view which doesn't contain any holes, at the cost of an incorrect reproduction of the depth which may lead to visible errors~\cite{Daribo2010}. Another way is the use of so-called inpainting techniques. Inpainting describes a process where holes in images are filled with synthesized content in a visually plausible manner, so that the viewer doesn't recognize that the content has been generated artificially. Recently, there has been quite extensive research on the adaptation of the inpainting algorithm by Criminisi et al.~\cite{Criminisi2004} for disocclusion filling. Criminisi's algorithm can be categorized into the group of so-called exemplar-based techniques, i.e., the algorithm uses patches of the image itself and copies these into the hole, thus exploiting the redundancy of natural images. Criminisi discovered that the order in which this filling process is executed determines the quality of the output image. He therefore introduced a confidence and a priority term with the intention to steer the filling process into the direction of isophotes, i.e., lines with constant luminance. However, using this algorithm directly for disocclusion filling in the context  of view synthesis leads to very poor results~\cite{Daribo2011}. Therefore, several modifications have been proposed. 

Oh et al.~\cite{Oh2009} explicitly modified the boundaries of the holes to only incorporate background pixels. Daribo and Saito~\cite{Daribo2011} proposed a depth-based modification to Criminisi's priority term to prioritize background pixels over foreground pixels. Gautier et al.~\cite{Gautier2011} replaced the color gradient in the priority term with a structure tensor based on the color of the texture and the structure of the depth map.

Criminisi's inpainting technique is a greedy algorithm, i.e., once a patch has been copied into the hole, it won't be changed regardless of the patches that follow in its neighbourhood. Komodakis and Tziritas~\cite{Komodakis2007} recognized this as a potential drawback and therefore introduced an inpainting algorithm based on the solution of a Markov random field. They demonstrate that this technique has the potential to significantly outperform the method of Criminisi in terms of visual quality of the inpainting result. In this contribution, we therefore propose an adaptation of the algorithm of Komodakis and Tziritas for disocclusion filling for view synthesis. The results compare favorably to the state of the art.
\begin{figure}[tb]
  \centering
   \includegraphics[height=3.0cm]{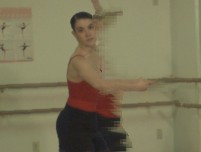}
  \caption{Unfortunately, Komodakis and Tziritas's algorithm cannot be used directly to fill disocclusions. The most obvious problem is bleeding of foreground objects into the background.}
  \label{fig:bleeding}
\end{figure}
We start by shortly reviewing the algorithm of Komodakis and Tziritas, introduce our extensions to make it applicable to view synthesis, show some of the results and compare it to the state of the art before concluding this paper.

\section{Algorithm}
\label{sec:algo}
\subsection{Komodakis and Tziritas's algorithm}
To make this paper self-contained, we will start by a brief summary of the algorithm of Komodakis and Tziritas so that we can introduce our extensions that make it applicable to view synthesis. However, for the sake of brevity, we would like to refer the reader to \cite{Komodakis2007} for further details. For clarity, we try to adapt the notation of \cite{Komodakis2007} as closely as possible. 

The hole-filling task is treated as a discrete, global optimization problem with a well-defined objective function. It therefore doesn't require any ad-hoc heuristics, such as the isophote continuation in Criminisi's algorithm, which may not be adequate in a general setting. First, we have to separate our image $I_{0}$ into a source region $S$ and a target region $T$, i.e., the hole(s) to be filled. In the view-synthesis problem, the location of the holes is determined by the scene geometry, and during the process of mapping the texture of the original view to the virtual view, we can simultaneously generate a mask which specifies the location of the holes. 
The results of this warping process can be seen in Figure~\ref{fig:warping}. 

\begin{figure}[tb]
\begin{subfigure}[t]{0.5\linewidth}%
    \centering
      \def\svgwidth{3.0cm}\input{img/mrf_overlay_new.pdf_tex}
    \caption{ }
    \label{fig:mrf_overlay}
\end{subfigure}\hfill
\begin{subfigure}[t]{0.5\linewidth}%
  \centering
    \setlength\figureheight{6cm}
    \setlength\figurewidth{6cm}
    \input{img/nodea.tex}
  \caption{ }
  \label{fig:nodea}
\end{subfigure}
\caption{Figure (a) shows a schematic visualization of the distribution of the nodes over the disocclusion. Nodes marked with yellow edges lie over foreground content and will be assigned a node potential $V_{p} = 0$. Figure (b) showcases the relative beliefs of a node at the border of the hole (Node A) and of any interior node (Node B) before the message passing step.}
\end{figure}
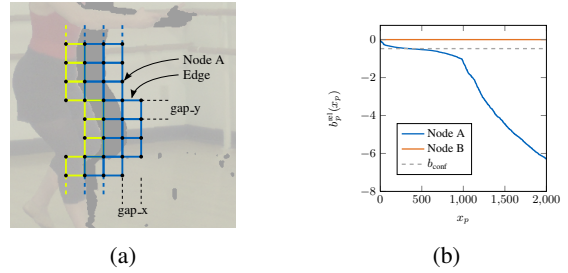

The image is then partitioned into small, overlapping patches of size $w \times h$ with a spacing of $\text{gap}_{x}$ and $\text{gap}_{y}$, respectively. Note that $\text{gap}_{x} < w$ and $\text{gap}_{y} < h$. The goal of the inpainting algorithm is then to find suitable patches from $S$ which can be filled into the holes $T$. To this end, Komodakis and Tziritas proposed a Markov network which consists of nodes $\nu$ at the positions of the patches inside and at the border of the hole. Each of these nodes has a set of labels associated with it which comprises candidate patches from $S$ to be inserted at the position of the node. The association of any of the labels to a node $p$ incurs specific costs which are defined as the node potential 
\begin{equation*}
V_{I,p}(x_{p}) = \!\!\!\!\!\!\!\!\!\!\!\!\!\!\! \sum_{dp\in \left [ -\frac{w}{2}\frac{w}{2} \right ]\times \left [ -\frac{h}{2}\frac{h}{2} \right ]} \!\!\!\!\!\!\!\!\!\!\!\!\!\!\! \mathcal{M}(p+dp)(I_{0}(p+dp)-I_{0}(x_{p}+dp))^{2}\text{,}
\end{equation*}
which describes how well the patch $x_{p}$ matches any available content from $S$, and a pairwise potential $V_{p,q}$, i.e., how well the patch matches the other patches in its 4-connected neighbourhood. $\mathcal{M}$ denotes a mask which is zero inside $T$, 1 else. The goal of the optimization problem is then to minimize the total energy of the MRF
\begin{equation*}
\mathcal{F}(\hat{x}) = \sum_{p\in\nu} V_{p}(\hat{x}_{p}) + \sum_{\left(p,q\right)\in\varepsilon} V_{pq}(\hat{x}_{p}, \hat{x}_{q})\text{,}
\end{equation*}
for which Komodakis and Tziritas proposed a \emph{priority-belief propagation} algorithm. In belief propagation, messages are exchanged along the edges $\varepsilon$ of connected nodes about the confidence in the association of a patch to a neighbouring node, which then in turn defines the belief $b_{p}(x_{p})$ each node has in its set of labels. As the number of possible patches is quite high in the setting of image completion, the computational cost of this BP-algorithm would be prohibitive. Komodakis and Tziritas therefore added a method called \emph{dynamic label pruning} based on priority. If a node has only a small set of labels in which it has a belief higher than a given confidence threshold, i.e., $b_{p}^{\text{rel}}(x_{p}) \geq b_{\text{conf}}$ with $b_{p}^{\text{rel}}(x_{p}) = b_{p}(x_{p})-b_{p}^{\text{max}}(x_{p})$, it will be assigned a high priority, that means it is quite confident about the assignment of its patch. On the other hand, if a node has similar beliefs in all of its labels, it may be considered indetermined and will be given a low priority. Nodes with high priority will be the ones to first get rid of all labels in which they have a low belief and then send efficient messages. Figure~\ref{fig:nodea} shows the distribution of the relative beliefs of a node with high priority which is usually located at the border of the hole. An interior node has a low priority as its node potential is zero and therefore has the same belief in all of its labels.

%
%
%
\subsection{Extensions for view synthesis}
Komodakis and Tziritas's algorithm is not directly applicable to the disocclusion problem in view synthesis as a naïve application leads to very poor results, as shown in Figure~\ref{fig:bleeding}. The most obvious problem is that there occurs bleeding of foreground objects into the background, which should be avoided. We therefore present our extensions which will deal with this problem.
As stated in the introduction, disocclusions occur at steep depth gradients, where there is a jump between a foreground object to the background of a scene. When we move the virtual camera to the right, the disocclusions will appear on the right side of foreground objects. We therefore adapt the idea of~\cite{Gautier2011} and others to steer the filling process into the opposite direction of the camera movement. In our setting, we achieve this by modifying the node potential of all nodes that are on the side of the disocclusion opposite to the camera movement, e.g., on the left side. These nodes, in Figure~\ref{fig:mrf_overlay} marked as yellow, are given a node potential $V_{p} = 0$. The algorithm thereby treats these just like interior nodes and they will get the lowest priority. As the MRF now doesn't have any support on the left side of the hole, the inpainting task has become somewhat similar to the texture synthesis task described in~\cite{Komodakis2007}. It is therefore necessary to introduce another term $V_{pq}^{0}(x_{p}, x_{q}) = w_{0}$ if $x_{p} - x_{q} \neq p - q$, $V_{pq}^{0}(x_{p}, x_{q}) =0$, else, to the cost function which enforces the coherence of the image by penalizing the filling of non-adjacent patches.

Furthermore, we modify the node potential 
\begin{equation*}
V_{p}(x_{p}) = V_{I,p}(x_{p}) + \lambda_{D} V_{D,p}(x_{p})
\end{equation*}
and the pairwise potential to not only accommodate for visual similarity between neighbouring nodes but also for similarity in depth. To this end, we add another term to both potentials which calculates the SSD 
\begin{equation*}
V_{D,p}(x_{p}) = \!\!\!\!\!\!\!\!\!\!\!\!\!\!\! \sum_{dp\in \left [ -\frac{w}{2}\frac{w}{2} \right ]\times \left [ -\frac{h}{2}\frac{h}{2} \right ]} \!\!\!\!\!\!\!\!\!\!\!\!\!\!\! \mathcal{M}(p+dp)(\mathcal{D}_{0}(p+dp)-\mathcal{D}_{0}(x_{p}+dp))^{2}
\end{equation*}
in the depth map $\mathcal{D}$, weighted by a factor $\lambda_{D}$. Thereby, we make sure that candidate patches are selected from similar depth ranges as the nodes which ensures consistency of the image and also improves the efficiency of the algorithm because it dramatically reduces the number of contemplable labels for each node.
%

\section{Evaluation}
\begin{figure*}[t]
\begin{subfigure}[b]{0.5\linewidth}
  \centering
  \includegraphics[width=.55\textwidth]{img/virtualimage}
  \caption{ }
\end{subfigure}\hfill
\begin{subfigure}[b]{.5\linewidth}
  \centering
  \includegraphics[width=.55\textwidth]{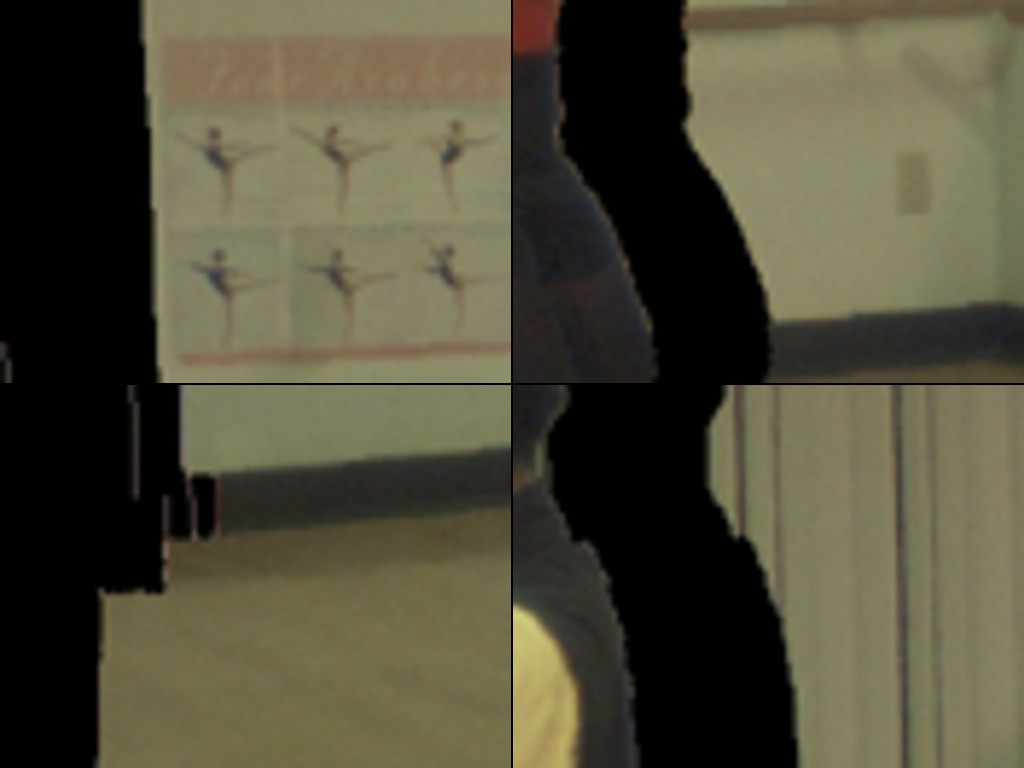}
  \caption{ }
\end{subfigure}\\
\begin{subfigure}[b]{0.5\linewidth}
  \centering
  \includegraphics[width=.55\textwidth]{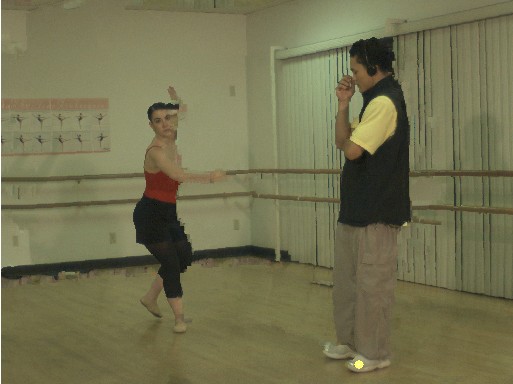}
  \caption{ }
\end{subfigure}\hfill
\begin{subfigure}[b]{.5\linewidth}
  \centering
  \includegraphics[width=.55\textwidth]{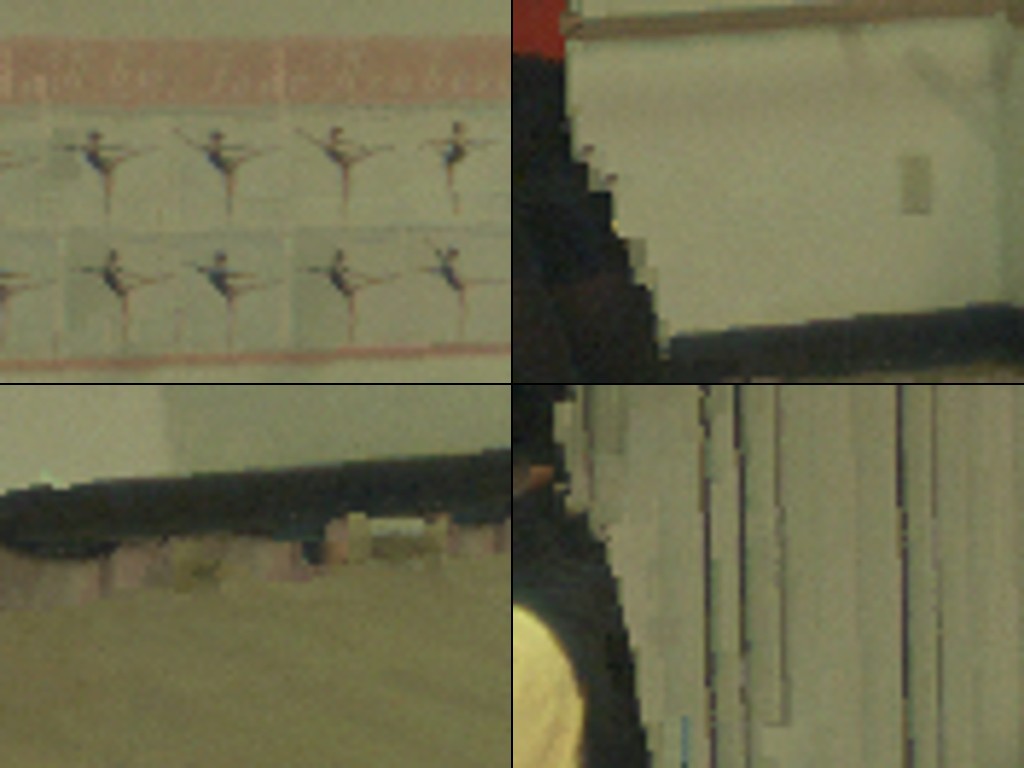}
  \caption{ }
\end{subfigure}\\
\begin{subfigure}[b]{0.5\linewidth}
  \centering
  \includegraphics[width=.55\textwidth]{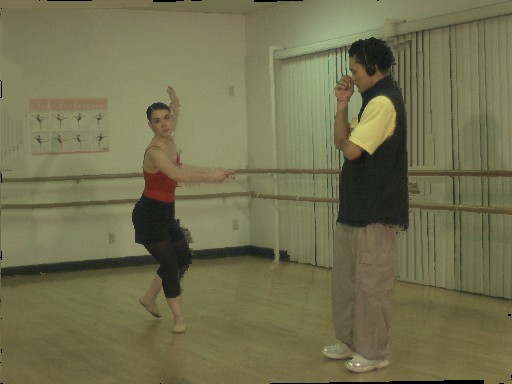}
  \caption{ }
\end{subfigure}\hfill
\begin{subfigure}[b]{.5\linewidth}
  \centering
  \includegraphics[width=.55\textwidth]{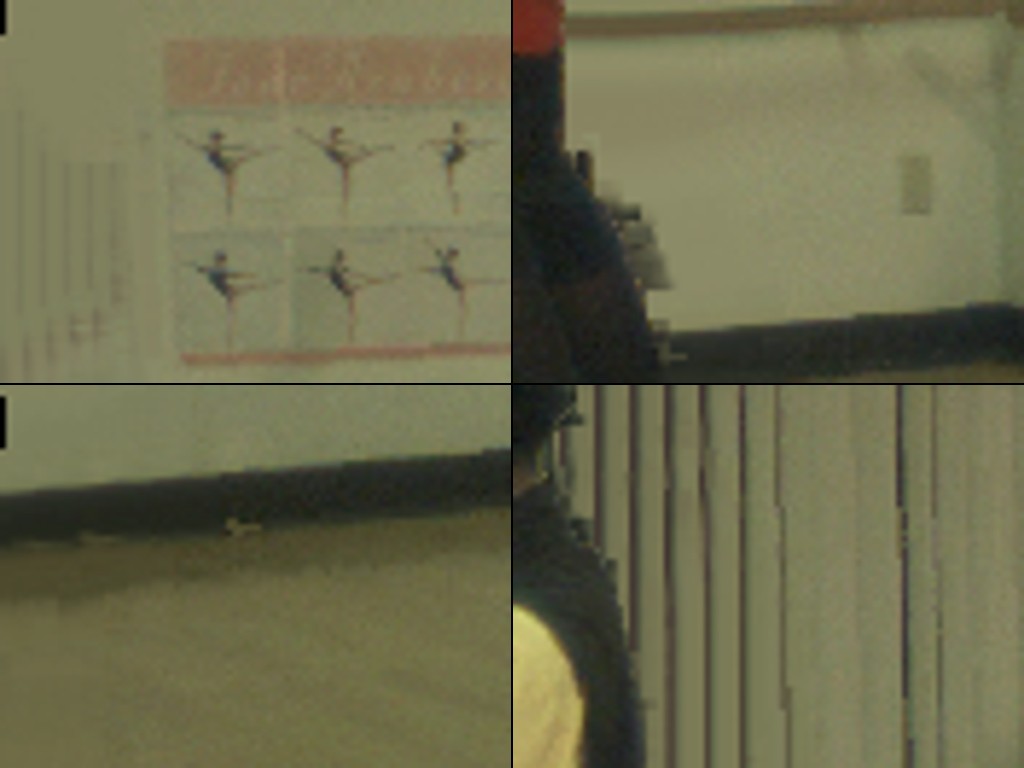}
  \caption{ }
\end{subfigure}\\
\begin{subfigure}[b]{0.5\linewidth}
  \centering
  \includegraphics[width=.55\textwidth]{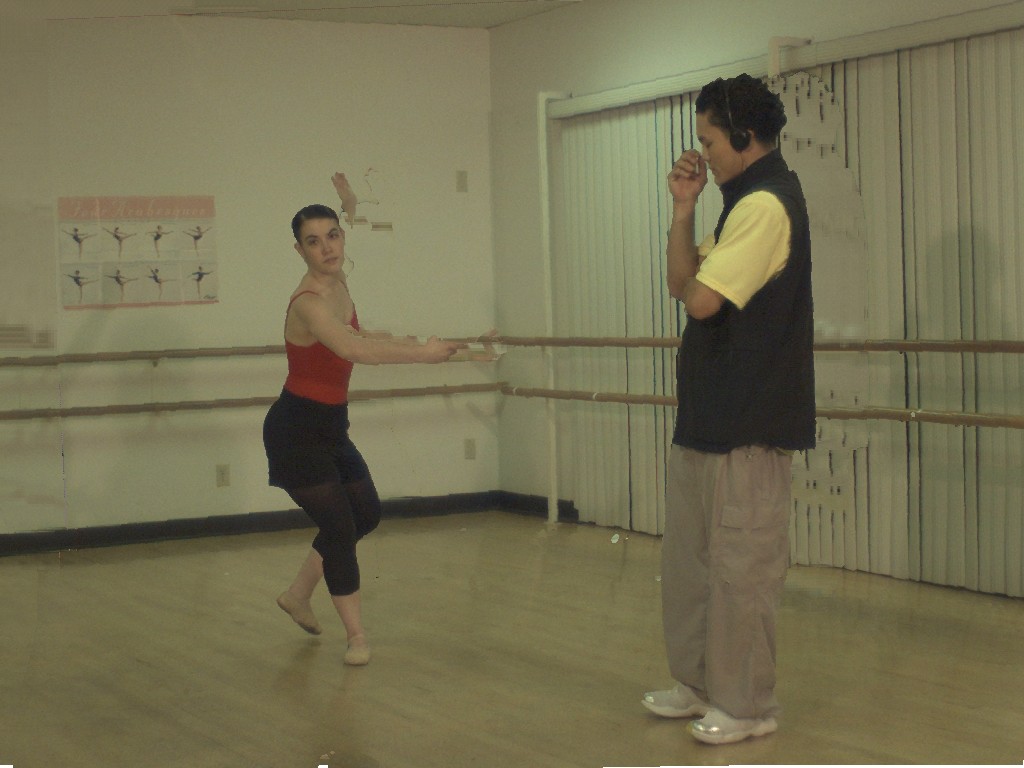}
  \caption{ }
\end{subfigure}\hfill
\begin{subfigure}[b]{.5\linewidth}
  \centering
  \includegraphics[width=.55\textwidth]{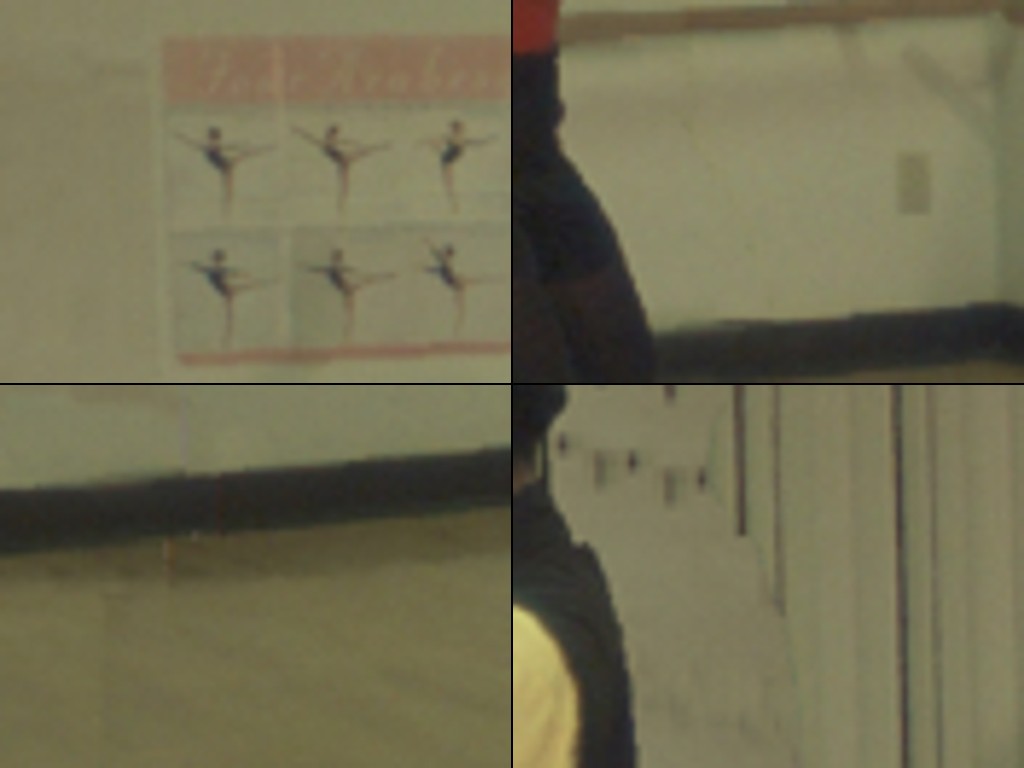}
  \caption{ }
\end{subfigure}\\
\begin{subfigure}[b]{0.5\linewidth}
  \centering
  \includegraphics[width=.55\textwidth]{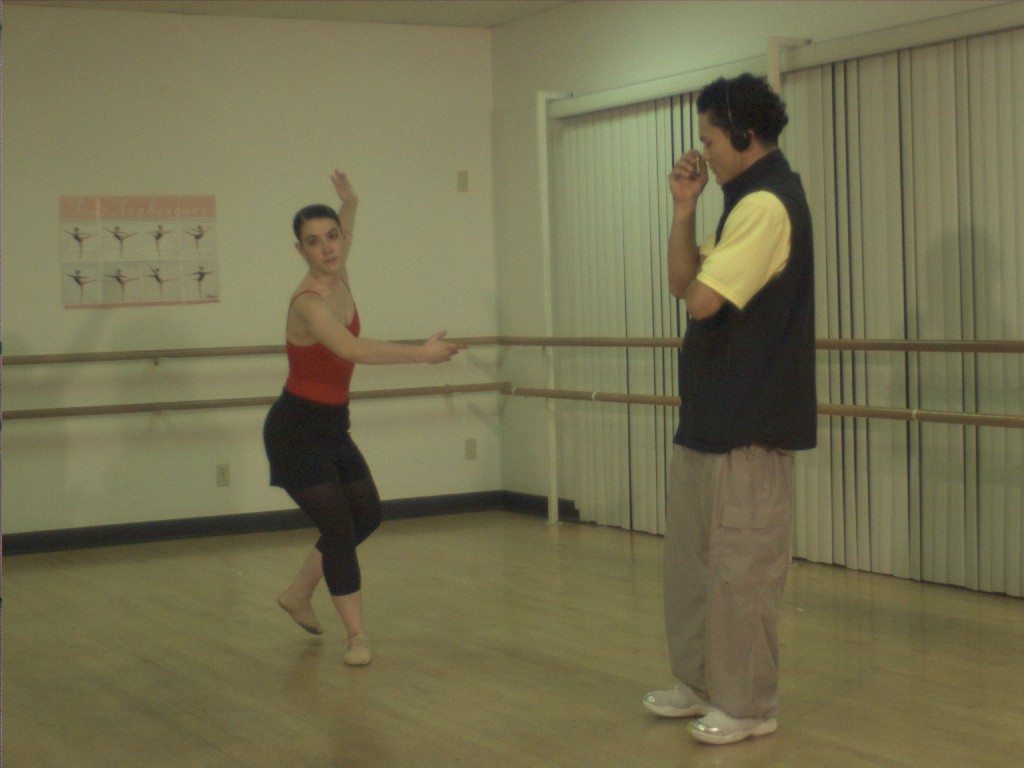}
  \caption{ }
\end{subfigure}\hfill
\begin{subfigure}[b]{.5\linewidth}
  \centering
  \includegraphics[width=.55\textwidth]{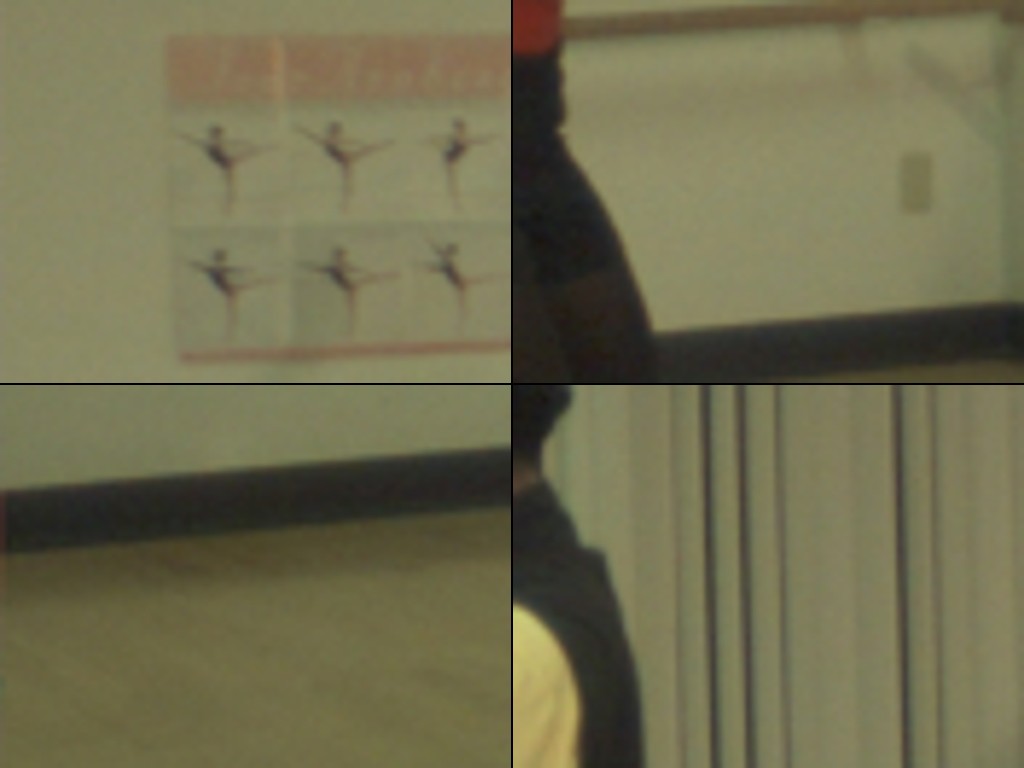}
  \caption{ }
\end{subfigure}
\caption{Inpainting results for the first frame of the \emph{Ballet} sequence of two state-of-the-art algorithms and our proposed algorithm. (a), (b) Synthesized View; (c), (d) Result of Daribo's method~\cite{Daribo2011}; (e), (f) Result of Gautier's method~\cite{Gautier2011}; (g), (h) Proposed method; (i), (j) Ground truth (Camera No.\ 4).}
\label{fig:res}
\end{figure*}

To evaluate the performance of our algorithm, we use the well-known Multiview Video-plus-Depth sequence \emph{Ballet} from Microsoft Research~\cite{Zitnick2004} because of its large baseline and because it allows us to make a fair comparison with two state-of-the-art algorithms~\cite{Daribo2011, Gautier2011}. %
For our evaluation we take the view from Camera No.\ 5 and create a virtual view which would be seen from Camera No.\ 4. We can therefore use the image from Camera No.\ 4 as a ground truth reference. We use the MPEG View Synthesis Reference Software (VSRS)~\cite{VSRS2008} in version 3.5 to generate the virtual view and use our algorithm to fill the disocclusions. As an objective measure for the quality of the inpainting result, we use SSIM. 
For completeness, we have also included the PSNR values, even though we think that PSNR is hardly a suitable measure to judge the quality of an inpainting algorithm. We also provide both measures for the regions which have been inpainted, only. The parameters of our algorithm have been chosen on the basis of the recommendations in \cite{Komodakis2007} and therefore weren't specifically tuned to the sequence; with the exception of the newly introduced parameter $\lambda_{D}$ which was set to $3$ to accommodate for the difference in the number of channels between the image and the depth map. The results can be seen in Table~\ref{tab:res} and in Figure~\ref{fig:res}.

\section{Conclusion \& Acknowledgements}
We have presented a new algorithm for disocclusion handling in view synthesis. A global optimization approach on Markov random fields incorporating the information of the depth map not only leads to consistent inpainting results but also to a higher algorithmic efficiency due to rigorous label pruning based on depth range. Objective evaluation on a standard dataset shows a significant improvement in image quality compared to the state of the art. 

We would like to thank Josselin Gautier for providing the software and results for his algorithm and the Microsoft Research team for providing the \emph{Ballet} data set.

\begin{table}[htb]
\caption{Objective evaluation of the inpainting result}
\begin{tabularx}{\linewidth}{|X|c|c|c|}
\hline 
 & \cite{Daribo2011} & \cite{Gautier2011} & proposed \\ 
\hline 
$\text{PSNR}_{Y}$ [dB] & 30.3 & 31.4 & \textbf{33.2} \\ 
\hline 
$\text{PSNR}_{Y}$ holes only [dB] & 24.2 & 24.0 & \textbf{26.2} \\ 
\hline 
$\text{SSIM}$ & 0.87 & 0.88 & \textbf{0.93} \\ 
\hline 
$\text{SSIM}$ holes only & 0.68 & 0.69 & \textbf{0.73} \\ 
\hline 
\end{tabularx}\label{tab:res}
\end{table}
%
\bibliographystyle{IEEEbib}
\bibliography{refs}
\end{document}

%% file: img/mrf_overlay_new.pdf_tex
\begingroup%
  \makeatletter%
  \providecommand\color[2][]{%
    \errmessage{(Inkscape) Color is used for the text in Inkscape, but the package 'color.sty' is not loaded}%
    \renewcommand\color[2][]{}%
  }%
  \providecommand\transparent[1]{%
    \errmessage{(Inkscape) Transparency is used (non-zero) for the text in Inkscape, but the package 'transparent.sty' is not loaded}%
    \renewcommand\transparent[1]{}%
  }%
  \providecommand\rotatebox[2]{#2}%
  \ifx\svgwidth\undefined%
    \setlength{\unitlength}{240bp}%
    \ifx\svgscale\undefined%
      \relax%
    \else%
      \setlength{\unitlength}{\unitlength * \real{\svgscale}}%
    \fi%
  \else%
    \setlength{\unitlength}{\svgwidth}%
  \fi%
  \global\let\svgwidth\undefined%
  \global\let\svgscale\undefined%
  \makeatother%
  \begin{picture}(1,1)%
    \put(0,0){\includegraphics[width=\unitlength]{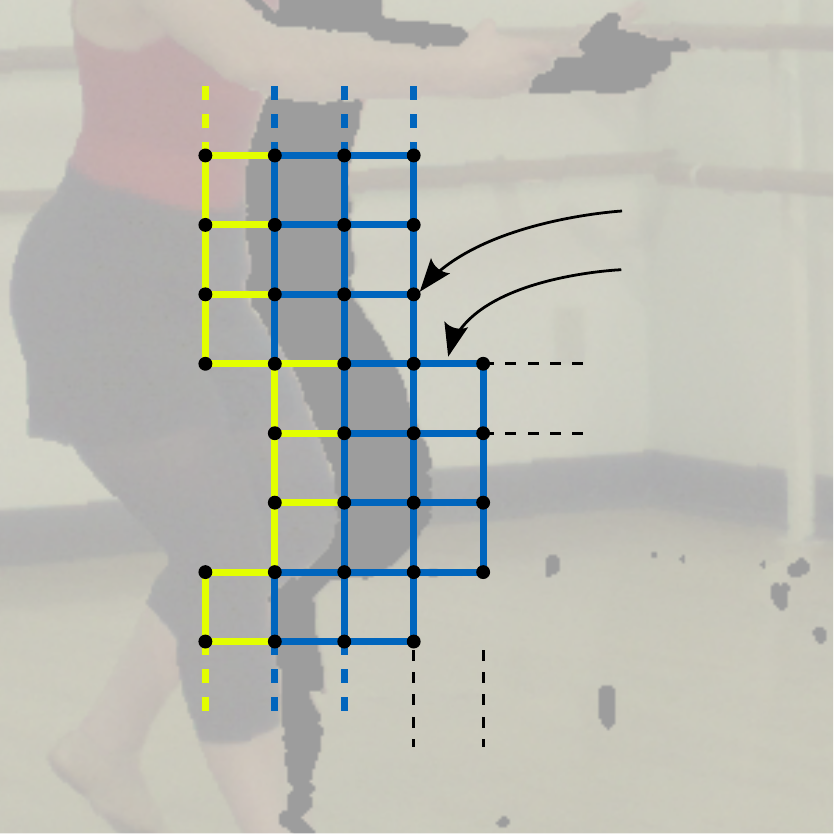}}%
    \put(0.70982108,0.512521){\color[rgb]{0,0,0}\makebox(0,0)[lb]{\tiny{\smash{gap\_y}}}}%
    \put(0.4764498,0.06013472){\color[rgb]{0,0,0}\makebox(0,0)[lb]{\tiny{\smash{gap\_x}}}}%
    \put(0.7664498,0.72683476){\color[rgb]{0,0,0}\makebox(0,0)[lb]{\tiny{\smash{Node A}}}}%
    \put(0.7664498,0.66016809){\color[rgb]{0,0,0}\makebox(0,0)[lb]{\tiny{\smash{Edge}}}}%
  \end{picture}%
\endgroup%

%% file: img/nodea.tex
%
%
\begin{tikzpicture}[scale=0.5]

\definecolor{mycolor1}{rgb}{0,0.396,0.741}
\definecolor{mycolor2}{rgb}{0.890,0.447,0.133}
\definecolor{mycolor3}{rgb}{0.65,0.65,0.65}

\begin{axis}[%
width=\figurewidth,
height=\figureheight,
xmin=0, xmax=2000,
ymin=-8, ymax=0.75,
xlabel={$x_p$},
ylabel={$b_{p}^{\text{rel}}(x_{p})$},
axis on top,
legend entries={Node A, Node B, $b_{\text{conf}}$},
legend style={at={(0.1,0.1)},anchor=south west,nodes=right}]
]
\addplot [
color=mycolor1,
solid,
line width=1.0pt
]
coordinates{
 (1,0)(2,-0.0360323)(3,-0.0451211)(4,-0.0826298)(5,-0.0829373)(6,-0.0863053)(7,-0.091088)(8,-0.0924875)(9,-0.0929027)(10,-0.0979162)(11,-0.10496)(12,-0.105021)(13,-0.111388)(14,-0.111865)(15,-0.115033)(16,-0.122368)(17,-0.122384)(18,-0.124506)(19,-0.130242)(20,-0.133272)(21,-0.143099)(22,-0.143729)(23,-0.152726)(24,-0.152726)(25,-0.155725)(26,-0.160923)(27,-0.161999)(28,-0.16303)(29,-0.166536)(30,-0.170704)(31,-0.172195)(32,-0.175625)(33,-0.179608)(34,-0.18173)(35,-0.195433)(36,-0.20975)(37,-0.211965)(38,-0.223299)(39,-0.224083)(40,-0.239846)(41,-0.24083)(42,-0.242015)(43,-0.255225)(44,-0.255379)(45,-0.25564)(46,-0.258824)(47,-0.259131)(48,-0.266421)(49,-0.267589)(50,-0.268727)(51,-0.269312)(52,-0.272834)(53,-0.27694)(54,-0.278677)(55,-0.278862)(56,-0.279616)(57,-0.283537)(58,-0.28406)(59,-0.287659)(60,-0.288612)(61,-0.288627)(62,-0.289227)(63,-0.292349)(64,-0.292995)(65,-0.295517)(66,-0.298824)(67,-0.300038)(68,-0.300331)(69,-0.3009)(70,-0.305529)(71,-0.305698)(72,-0.30722)(73,-0.307543)(74,-0.308497)(75,-0.308973)(76,-0.309081)(77,-0.309435)(78,-0.310588)(79,-0.311219)(80,-0.311588)(81,-0.311895)(82,-0.312341)(83,-0.313033)(84,-0.313602)(85,-0.314494)(86,-0.314787)(87,-0.315033)(88,-0.315402)(89,-0.316032)(90,-0.321738)(91,-0.322768)(92,-0.32306)(93,-0.323737)(94,-0.325183)(95,-0.326459)(96,-0.32729)(97,-0.327797)(98,-0.328181)(99,-0.329519)(100,-0.330042)(101,-0.330396)(102,-0.331027)(103,-0.33118)(104,-0.331765)(105,-0.331995)(106,-0.332303)(107,-0.333764)(108,-0.334287)(109,-0.335179)(110,-0.339454)(111,-0.339792)(112,-0.341392)(113,-0.34213)(114,-0.342376)(115,-0.342637)(116,-0.34479)(117,-0.347682)(118,-0.347759)(119,-0.348035)(120,-0.348558)(121,-0.349173)(122,-0.349666)(123,-0.349865)(124,-0.350711)(125,-0.351726)(126,-0.351757)(127,-0.354402)(128,-0.355556)(129,-0.356647)(130,-0.357293)(131,-0.358062)(132,-0.358601)(133,-0.35877)(134,-0.361615)(135,-0.361676)(136,-0.36183)(137,-0.36243)(138,-0.363291)(139,-0.363691)(140,-0.363799)(141,-0.363937)(142,-0.364798)(143,-0.366674)(144,-0.366828)(145,-0.368366)(146,-0.369781)(147,-0.370012)(148,-0.371057)(149,-0.371749)(150,-0.372195)(151,-0.373195)(152,-0.375163)(153,-0.377409)(154,-0.377624)(155,-0.378593)(156,-0.378624)(157,-0.37907)(158,-0.379931)(159,-0.382099)(160,-0.383714)(161,-0.384006)(162,-0.384468)(163,-0.385436)(164,-0.388097)(165,-0.388989)(166,-0.390327)(167,-0.390404)(168,-0.390788)(169,-0.390819)(170,-0.392018)(171,-0.393433)(172,-0.393679)(173,-0.393772)(174,-0.394125)(175,-0.394187)(176,-0.395048)(177,-0.396032)(178,-0.397586)(179,-0.397955)(180,-0.399216)(181,-0.400446)(182,-0.400661)(183,-0.401399)(184,-0.401753)(185,-0.402168)(186,-0.402476)(187,-0.402661)(188,-0.402891)(189,-0.403799)(190,-0.405459)(191,-0.405875)(192,-0.407259)(193,-0.407951)(194,-0.408566)(195,-0.410427)(196,-0.41098)(197,-0.411365)(198,-0.411672)(199,-0.412303)(200,-0.412364)(201,-0.41278)(202,-0.414671)(203,-0.414748)(204,-0.415394)(205,-0.415855)(206,-0.416486)(207,-0.417301)(208,-0.417439)(209,-0.417455)(210,-0.41807)(211,-0.418716)(212,-0.4191)(213,-0.419146)(214,-0.419393)(215,-0.419777)(216,-0.420161)(217,-0.422945)(218,-0.424529)(219,-0.424606)(220,-0.424698)(221,-0.424729)(222,-0.426359)(223,-0.426436)(224,-0.426544)(225,-0.427051)(226,-0.427051)(227,-0.427205)(228,-0.427436)(229,-0.428112)(230,-0.428481)(231,-0.428897)(232,-0.429235)(233,-0.429958)(234,-0.430173)(235,-0.430327)(236,-0.430788)(237,-0.430788)(238,-0.432218)(239,-0.432388)(240,-0.432587)(241,-0.433049)(242,-0.433172)(243,-0.433233)(244,-0.433233)(245,-0.434002)(246,-0.434079)(247,-0.434171)(248,-0.435002)(249,-0.436478)(250,-0.436678)(251,-0.436955)(252,-0.437216)(253,-0.437586)(254,-0.438893)(255,-0.439093)(256,-0.439123)(257,-0.43917)(258,-0.440631)(259,-0.440892)(260,-0.441922)(261,-0.442276)(262,-0.442291)(263,-0.442414)(264,-0.444183)(265,-0.44529)(266,-0.44569)(267,-0.447166)(268,-0.447428)(269,-0.447597)(270,-0.448058)(271,-0.448827)(272,-0.449842)(273,-0.450273)(274,-0.450673)(275,-0.450765)(276,-0.45158)(277,-0.451596)(278,-0.451903)(279,-0.452072)(280,-0.452441)(281,-0.452626)(282,-0.453918)(283,-0.454164)(284,-0.454179)(285,-0.454533)(286,-0.454764)(287,-0.455363)(288,-0.455671)(289,-0.455748)(290,-0.455932)(291,-0.456963)(292,-0.45704)(293,-0.457809)(294,-0.458331)(295,-0.458424)(296,-0.458531)(297,-0.458593)(298,-0.459146)(299,-0.459623)(300,-0.459962)(301,-0.460531)(302,-0.460854)(303,-0.461207)(304,-0.461407)(305,-0.461607)(306,-0.462115)(307,-0.462268)(308,-0.462499)(309,-0.462561)(310,-0.462576)(311,-0.46273)(312,-0.46276)(313,-0.463222)(314,-0.46396)(315,-0.464022)(316,-0.464529)(317,-0.464667)(318,-0.464837)(319,-0.465037)(320,-0.465313)(321,-0.465544)(322,-0.465867)(323,-0.466298)(324,-0.466544)(325,-0.468358)(326,-0.468789)(327,-0.469773)(328,-0.469804)(329,-0.469927)(330,-0.470373)(331,-0.470865)(332,-0.47128)(333,-0.471649)(334,-0.471803)(335,-0.471895)(336,-0.471957)(337,-0.471957)(338,-0.472034)(339,-0.472034)(340,-0.472065)(341,-0.472203)(342,-0.47308)(343,-0.47328)(344,-0.473787)(345,-0.473802)(346,-0.475433)(347,-0.475494)(348,-0.475879)(349,-0.477109)(350,-0.477109)(351,-0.477186)(352,-0.477878)(353,-0.478401)(354,-0.478401)(355,-0.478447)(356,-0.478447)(357,-0.478616)(358,-0.478847)(359,-0.479523)(360,-0.479677)(361,-0.480123)(362,-0.480384)(363,-0.481246)(364,-0.481292)(365,-0.481769)(366,-0.481799)(367,-0.482184)(368,-0.482368)(369,-0.482584)(370,-0.482599)(371,-0.483076)(372,-0.483137)(373,-0.483583)(374,-0.48366)(375,-0.483968)(376,-0.484168)(377,-0.484291)(378,-0.484998)(379,-0.485921)(380,-0.486305)(381,-0.486367)(382,-0.487166)(383,-0.487582)(384,-0.487874)(385,-0.488489)(386,-0.488627)(387,-0.488812)(388,-0.489366)(389,-0.489566)(390,-0.489596)(391,-0.489904)(392,-0.489965)(393,-0.490119)(394,-0.490196)(395,-0.490458)(396,-0.490565)(397,-0.491473)(398,-0.491826)(399,-0.492257)(400,-0.49278)(401,-0.492933)(402,-0.49318)(403,-0.493887)(404,-0.494579)(405,-0.495056)(406,-0.495194)(407,-0.495532)(408,-0.495625)(409,-0.495702)(410,-0.495748)(411,-0.495917)(412,-0.496655)(413,-0.496701)(414,-0.496732)(415,-0.497255)(416,-0.497286)(417,-0.497532)(418,-0.497655)(419,-0.498085)(420,-0.498101)(421,-0.498162)(422,-0.498224)(423,-0.498824)(424,-0.498962)(425,-0.499085)(426,-0.499285)(427,-0.499639)(428,-0.4997)(429,-0.501392)(430,-0.501423)(431,-0.501484)(432,-0.501592)(433,-0.501761)(434,-0.501915)(435,-0.502438)(436,-0.502484)(437,-0.50273)(438,-0.503237)(439,-0.503529)(440,-0.50376)(441,-0.503806)(442,-0.503975)(443,-0.504221)(444,-0.504606)(445,-0.504867)(446,-0.504975)(447,-0.505144)(448,-0.505544)(449,-0.505559)(450,-0.505759)(451,-0.50722)(452,-0.507282)(453,-0.507835)(454,-0.508481)(455,-0.508527)(456,-0.509066)(457,-0.509343)(458,-0.509558)(459,-0.511096)(460,-0.511526)(461,-0.512495)(462,-0.513095)(463,-0.513987)(464,-0.514018)(465,-0.514095)(466,-0.514264)(467,-0.514387)(468,-0.514771)(469,-0.515325)(470,-0.515602)(471,-0.515879)(472,-0.517078)(473,-0.517555)(474,-0.517678)(475,-0.518001)(476,-0.518124)(477,-0.518585)(478,-0.518647)(479,-0.518831)(480,-0.519477)(481,-0.519646)(482,-0.519908)(483,-0.520431)(484,-0.521092)(485,-0.522614)(486,-0.522753)(487,-0.522953)(488,-0.523322)(489,-0.524045)(490,-0.524229)(491,-0.524383)(492,-0.524506)(493,-0.524767)(494,-0.52489)(495,-0.525198)(496,-0.526305)(497,-0.526305)(498,-0.52689)(499,-0.52689)(500,-0.527336)(501,-0.527951)(502,-0.528935)(503,-0.528981)(504,-0.529412)(505,-0.529689)(506,-0.530719)(507,-0.53098)(508,-0.531042)(509,-0.531103)(510,-0.531196)(511,-0.531242)(512,-0.531273)(513,-0.531519)(514,-0.531857)(515,-0.531934)(516,-0.531949)(517,-0.532057)(518,-0.532334)(519,-0.532364)(520,-0.533564)(521,-0.534148)(522,-0.534271)(523,-0.534287)(524,-0.534348)(525,-0.534364)(526,-0.534425)(527,-0.534671)(528,-0.534702)(529,-0.535163)(530,-0.53544)(531,-0.537116)(532,-0.537286)(533,-0.537993)(534,-0.538454)(535,-0.538501)(536,-0.538608)(537,-0.538931)(538,-0.539531)(539,-0.539531)(540,-0.539546)(541,-0.539746)(542,-0.539777)(543,-0.5403)(544,-0.540315)(545,-0.540423)(546,-0.5407)(547,-0.541192)(548,-0.541607)(549,-0.541869)(550,-0.542222)(551,-0.544022)(552,-0.544468)(553,-0.544529)(554,-0.544621)(555,-0.54496)(556,-0.545067)(557,-0.545175)(558,-0.545759)(559,-0.545836)(560,-0.546436)(561,-0.547389)(562,-0.548281)(563,-0.549542)(564,-0.549758)(565,-0.549865)(566,-0.550142)(567,-0.550296)(568,-0.550434)(569,-0.551357)(570,-0.551742)(571,-0.551911)(572,-0.552203)(573,-0.552418)(574,-0.553049)(575,-0.55328)(576,-0.553356)(577,-0.553372)(578,-0.553818)(579,-0.55411)(580,-0.555279)(581,-0.555755)(582,-0.555771)(583,-0.55634)(584,-0.556524)(585,-0.557616)(586,-0.558339)(587,-0.559047)(588,-0.559616)(589,-0.559846)(590,-0.560062)(591,-0.560569)(592,-0.560569)(593,-0.5606)(594,-0.560907)(595,-0.561599)(596,-0.562553)(597,-0.564798)(598,-0.565167)(599,-0.565167)(600,-0.56529)(601,-0.565367)(602,-0.565736)(603,-0.566228)(604,-0.567536)(605,-0.567705)(606,-0.568043)(607,-0.568797)(608,-0.56915)(609,-0.570135)(610,-0.570365)(611,-0.57115)(612,-0.571503)(613,-0.571611)(614,-0.571811)(615,-0.572672)(616,-0.573118)(617,-0.574548)(618,-0.57501)(619,-0.575087)(620,-0.575148)(621,-0.576701)(622,-0.577009)(623,-0.577885)(624,-0.578178)(625,-0.578516)(626,-0.579992)(627,-0.580054)(628,-0.580192)(629,-0.581392)(630,-0.581561)(631,-0.582176)(632,-0.583837)(633,-0.58416)(634,-0.584852)(635,-0.585421)(636,-0.586313)(637,-0.586544)(638,-0.588481)(639,-0.589466)(640,-0.590081)(641,-0.590158)(642,-0.590865)(643,-0.591895)(644,-0.592987)(645,-0.593264)(646,-0.593479)(647,-0.594648)(648,-0.596863)(649,-0.597155)(650,-0.597309)(651,-0.597985)(652,-0.59817)(653,-0.5992)(654,-0.599662)(655,-0.600384)(656,-0.602045)(657,-0.602384)(658,-0.603168)(659,-0.603629)(660,-0.604014)(661,-0.604075)(662,-0.604091)(663,-0.604552)(664,-0.60466)(665,-0.604906)(666,-0.606521)(667,-0.60689)(668,-0.607828)(669,-0.608966)(670,-0.609089)(671,-0.609258)(672,-0.610242)(673,-0.611457)(674,-0.612395)(675,-0.612964)(676,-0.613795)(677,-0.614533)(678,-0.615333)(679,-0.615963)(680,-0.616471)(681,-0.617178)(682,-0.617439)(683,-0.617824)(684,-0.618501)(685,-0.619562)(686,-0.619592)(687,-0.62113)(688,-0.621945)(689,-0.622868)(690,-0.623083)(691,-0.626805)(692,-0.627051)(693,-0.627974)(694,-0.631726)(695,-0.632418)(696,-0.632434)(697,-0.636324)(698,-0.636555)(699,-0.638216)(700,-0.640769)(701,-0.640861)(702,-0.641261)(703,-0.642691)(704,-0.648428)(705,-0.649781)(706,-0.650565)(707,-0.650796)(708,-0.653041)(709,-0.653518)(710,-0.653702)(711,-0.656794)(712,-0.656901)(713,-0.658639)(714,-0.659869)(715,-0.660438)(716,-0.66133)(717,-0.661638)(718,-0.662176)(719,-0.663899)(720,-0.664391)(721,-0.664575)(722,-0.666128)(723,-0.666605)(724,-0.667067)(725,-0.667497)(726,-0.667589)(727,-0.669358)(728,-0.669681)(729,-0.669835)(730,-0.669973)(731,-0.671373)(732,-0.671819)(733,-0.671911)(734,-0.673818)(735,-0.674187)(736,-0.67511)(737,-0.675894)(738,-0.678124)(739,-0.679077)(740,-0.680584)(741,-0.68243)(742,-0.68303)(743,-0.684752)(744,-0.685598)(745,-0.687028)(746,-0.687797)(747,-0.688781)(748,-0.693472)(749,-0.693964)(750,-0.694241)(751,-0.694933)(752,-0.695502)(753,-0.695686)(754,-0.695948)(755,-0.696547)(756,-0.696917)(757,-0.697962)(758,-0.698547)(759,-0.699023)(760,-0.69927)(761,-0.699577)(762,-0.700561)(763,-0.701176)(764,-0.701238)(765,-0.702053)(766,-0.702807)(767,-0.703268)(768,-0.704052)(769,-0.704683)(770,-0.705021)(771,-0.706328)(772,-0.706467)(773,-0.706867)(774,-0.70925)(775,-0.709819)(776,-0.712449)(777,-0.714048)(778,-0.717647)(779,-0.718693)(780,-0.7188)(781,-0.720323)(782,-0.720477)(783,-0.721845)(784,-0.722368)(785,-0.722414)(786,-0.722768)(787,-0.724091)(788,-0.72466)(789,-0.724952)(790,-0.726644)(791,-0.728228)(792,-0.728504)(793,-0.728874)(794,-0.731642)(795,-0.732933)(796,-0.734256)(797,-0.73684)(798,-0.739485)(799,-0.739915)(800,-0.741715)(801,-0.741745)(802,-0.743852)(803,-0.744944)(804,-0.745206)(805,-0.746405)(806,-0.746528)(807,-0.746544)(808,-0.752295)(809,-0.75308)(810,-0.754171)(811,-0.755771)(812,-0.756278)(813,-0.756755)(814,-0.762522)(815,-0.762553)(816,-0.76306)(817,-0.763306)(818,-0.763629)(819,-0.766213)(820,-0.769012)(821,-0.770181)(822,-0.770196)(823,-0.770334)(824,-0.770873)(825,-0.772949)(826,-0.774887)(827,-0.777547)(828,-0.777855)(829,-0.778439)(830,-0.781869)(831,-0.782576)(832,-0.784391)(833,-0.784606)(834,-0.787851)(835,-0.78945)(836,-0.791419)(837,-0.791542)(838,-0.794571)(839,-0.794602)(840,-0.795233)(841,-0.79534)(842,-0.798401)(843,-0.799385)(844,-0.801738)(845,-0.802537)(846,-0.802707)(847,-0.802799)(848,-0.803168)(849,-0.804121)(850,-0.808274)(851,-0.809181)(852,-0.809812)(853,-0.810304)(854,-0.81158)(855,-0.812149)(856,-0.812441)(857,-0.814148)(858,-0.814394)(859,-0.81604)(860,-0.816424)(861,-0.81684)(862,-0.816886)(863,-0.818116)(864,-0.819885)(865,-0.820961)(866,-0.823483)(867,-0.82579)(868,-0.827605)(869,-0.828589)(870,-0.831803)(871,-0.832003)(872,-0.834325)(873,-0.834387)(874,-0.835402)(875,-0.836094)(876,-0.836524)(877,-0.841461)(878,-0.846767)(879,-0.847951)(880,-0.848781)(881,-0.851073)(882,-0.851519)(883,-0.85481)(884,-0.85484)(885,-0.856286)(886,-0.85767)(887,-0.85827)(888,-0.859469)(889,-0.860884)(890,-0.863714)(891,-0.86376)(892,-0.863883)(893,-0.865098)(894,-0.865928)(895,-0.870681)(896,-0.87085)(897,-0.872065)(898,-0.872849)(899,-0.873326)(900,-0.874525)(901,-0.874756)(902,-0.8806)(903,-0.880738)(904,-0.881784)(905,-0.88346)(906,-0.883537)(907,-0.88509)(908,-0.888643)(909,-0.889627)(910,-0.891365)(911,-0.892534)(912,-0.892657)(913,-0.894533)(914,-0.89501)(915,-0.896071)(916,-0.898685)(917,-0.898947)(918,-0.899039)(919,-0.899746)(920,-0.902299)(921,-0.903868)(922,-0.908143)(923,-0.908789)(924,-0.912157)(925,-0.915663)(926,-0.91614)(927,-0.919539)(928,-0.922799)(929,-0.924983)(930,-0.933487)(931,-0.933918)(932,-0.939193)(933,-0.9411)(934,-0.943022)(935,-0.943929)(936,-0.94619)(937,-0.94822)(938,-0.948835)(939,-0.950696)(940,-0.953664)(941,-0.954633)(942,-0.954833)(943,-0.956463)(944,-0.958862)(945,-0.959216)(946,-0.959908)(947,-0.960661)(948,-0.96649)(949,-0.968566)(950,-0.969196)(951,-0.975702)(952,-0.976547)(953,-0.97684)(954,-0.977655)(955,-0.978577)(956,-0.9795)(957,-0.985021)(958,-0.985221)(959,-0.98639)(960,-0.989004)(961,-0.989435)(962,-0.989558)(963,-0.990681)(964,-0.991649)(965,-0.993326)(966,-0.994125)(967,-0.997463)(968,-0.998308)(969,-0.999431)(970,-1.00349)(971,-1.004)(972,-1.0052)(973,-1.00591)(974,-1.00674)(975,-1.00737)(976,-1.0111)(977,-1.01127)(978,-1.01641)(979,-1.01845)(980,-1.01849)(981,-1.01999)(982,-1.0282)(983,-1.02833)(984,-1.02956)(985,-1.03589)(986,-1.04285)(987,-1.05573)(988,-1.08089)(989,-1.09333)(990,-1.0939)(991,-1.09489)(992,-1.10222)(993,-1.10845)(994,-1.1307)(995,-1.13283)(996,-1.14927)(997,-1.16857)(998,-1.1716)(999,-1.18794)(1000,-1.20163)(1001,-1.21097)(1002,-1.22198)(1003,-1.24864)(1004,-1.25063)(1005,-1.26155)(1006,-1.27045)(1007,-1.2727)(1008,-1.28543)(1009,-1.28824)(1010,-1.29464)(1011,-1.29932)(1012,-1.30279)(1013,-1.30662)(1014,-1.31274)(1015,-1.32824)(1016,-1.34033)(1017,-1.35489)(1018,-1.35699)(1019,-1.35708)(1020,-1.36234)(1021,-1.37307)(1022,-1.37579)(1023,-1.39071)(1024,-1.40458)(1025,-1.40897)(1026,-1.42296)(1027,-1.44335)(1028,-1.45073)(1029,-1.48324)(1030,-1.48852)(1031,-1.49226)(1032,-1.50028)(1033,-1.5009)(1034,-1.53212)(1035,-1.54247)(1036,-1.54587)(1037,-1.55343)(1038,-1.5666)(1039,-1.56917)(1040,-1.57542)(1041,-1.58201)(1042,-1.58954)(1043,-1.59091)(1044,-1.5934)(1045,-1.59354)(1046,-1.62954)(1047,-1.6306)(1048,-1.63466)(1049,-1.64201)(1050,-1.68466)(1051,-1.68691)(1052,-1.6981)(1053,-1.70158)(1054,-1.71106)(1055,-1.71256)(1056,-1.71892)(1057,-1.72709)(1058,-1.73858)(1059,-1.7515)(1060,-1.761)(1061,-1.7676)(1062,-1.77772)(1063,-1.77967)(1064,-1.78267)(1065,-1.78548)(1066,-1.78787)(1067,-1.78907)(1068,-1.79074)(1069,-1.79646)(1070,-1.80481)(1071,-1.82582)(1072,-1.8321)(1073,-1.83239)(1074,-1.83393)(1075,-1.83791)(1076,-1.84154)(1077,-1.84438)(1078,-1.84531)(1079,-1.84663)(1080,-1.849)(1081,-1.8541)(1082,-1.8569)(1083,-1.85693)(1084,-1.86145)(1085,-1.86239)(1086,-1.86691)(1087,-1.87031)(1088,-1.871)(1089,-1.87508)(1090,-1.88929)(1091,-1.89498)(1092,-1.90757)(1093,-1.9092)(1094,-1.91043)(1095,-1.92074)(1096,-1.92135)(1097,-1.94228)(1098,-1.94633)(1099,-1.95166)(1100,-1.96183)(1101,-1.96194)(1102,-1.96961)(1103,-1.9699)(1104,-1.9705)(1105,-1.98484)(1106,-1.98764)(1107,-1.99199)(1108,-1.99482)(1109,-1.99552)(1110,-1.99743)(1111,-1.99952)(1112,-2.00826)(1113,-2.01063)(1114,-2.0137)(1115,-2.02362)(1116,-2.02867)(1117,-2.03548)(1118,-2.03772)(1119,-2.04291)(1120,-2.04667)(1121,-2.06101)(1122,-2.06659)(1123,-2.07174)(1124,-2.07932)(1125,-2.08191)(1126,-2.08561)(1127,-2.09046)(1128,-2.09861)(1129,-2.10441)(1130,-2.1095)(1131,-2.1111)(1132,-2.1174)(1133,-2.12077)(1134,-2.12291)(1135,-2.13298)(1136,-2.13502)(1137,-2.15709)(1138,-2.16406)(1139,-2.18052)(1140,-2.20448)(1141,-2.20669)(1142,-2.2114)(1143,-2.22319)(1144,-2.24549)(1145,-2.24644)(1146,-2.25695)(1147,-2.27857)(1148,-2.28424)(1149,-2.29169)(1150,-2.2987)(1151,-2.29972)(1152,-2.30084)(1153,-2.30379)(1154,-2.30624)(1155,-2.33398)(1156,-2.34378)(1157,-2.35656)(1158,-2.36997)(1159,-2.38178)(1160,-2.38356)(1161,-2.39431)(1162,-2.40963)(1163,-2.43846)(1164,-2.4424)(1165,-2.45213)(1166,-2.45539)(1167,-2.47067)(1168,-2.49015)(1169,-2.50156)(1170,-2.50215)(1171,-2.50393)(1172,-2.51339)(1173,-2.51453)(1174,-2.5164)(1175,-2.52243)(1176,-2.52921)(1177,-2.53701)(1178,-2.53818)(1179,-2.54321)(1180,-2.553)(1181,-2.58153)(1182,-2.59143)(1183,-2.592)(1184,-2.6038)(1185,-2.62654)(1186,-2.62714)(1187,-2.62784)(1188,-2.63546)(1189,-2.65463)(1190,-2.66508)(1191,-2.66564)(1192,-2.66659)(1193,-2.67053)(1194,-2.69942)(1195,-2.70333)(1196,-2.70456)(1197,-2.71031)(1198,-2.71552)(1199,-2.72129)(1200,-2.75251)(1201,-2.75768)(1202,-2.75771)(1203,-2.76691)(1204,-2.78484)(1205,-2.78581)(1206,-2.80029)(1207,-2.81072)(1208,-2.82095)(1209,-2.82439)(1210,-2.85084)(1211,-2.85556)(1212,-2.86258)(1213,-2.86816)(1214,-2.87825)(1215,-2.88137)(1216,-2.89893)(1217,-2.90022)(1218,-2.90318)(1219,-2.91402)(1220,-2.91732)(1221,-2.92551)(1222,-2.94211)(1223,-2.9459)(1224,-2.95216)(1225,-2.96503)(1226,-2.96718)(1227,-2.97295)(1228,-2.99574)(1229,-3.02728)(1230,-3.03059)(1231,-3.03568)(1232,-3.04378)(1233,-3.04386)(1234,-3.04394)(1235,-3.04604)(1236,-3.04663)(1237,-3.04798)(1238,-3.04858)(1239,-3.05241)(1240,-3.06491)(1241,-3.08458)(1242,-3.09058)(1243,-3.09533)(1244,-3.10124)(1245,-3.10544)(1246,-3.10556)(1247,-3.11691)(1248,-3.12052)(1249,-3.1206)(1250,-3.1262)(1251,-3.1275)(1252,-3.13303)(1253,-3.13956)(1254,-3.15302)(1255,-3.16829)(1256,-3.17555)(1257,-3.18908)(1258,-3.18923)(1259,-3.19163)(1260,-3.19719)(1261,-3.19759)(1262,-3.20486)(1263,-3.20915)(1264,-3.22514)(1265,-3.227)(1266,-3.23036)(1267,-3.2318)(1268,-3.23525)(1269,-3.2547)(1270,-3.26039)(1271,-3.26111)(1272,-3.26674)(1273,-3.27945)(1274,-3.28155)(1275,-3.28537)(1276,-3.28714)(1277,-3.29489)(1278,-3.29535)(1279,-3.30525)(1280,-3.31895)(1281,-3.32274)(1282,-3.32574)(1283,-3.33436)(1284,-3.33475)(1285,-3.33613)(1286,-3.34439)(1287,-3.36701)(1288,-3.37089)(1289,-3.38424)(1290,-3.38737)(1291,-3.38956)(1292,-3.39914)(1293,-3.40032)(1294,-3.40667)(1295,-3.41527)(1296,-3.41595)(1297,-3.41632)(1298,-3.41646)(1299,-3.42565)(1300,-3.42574)(1301,-3.42862)(1302,-3.44315)(1303,-3.44692)(1304,-3.44935)(1305,-3.45481)(1306,-3.45875)(1307,-3.46322)(1308,-3.46616)(1309,-3.46859)(1310,-3.47259)(1311,-3.47554)(1312,-3.47972)(1313,-3.48051)(1314,-3.50142)(1315,-3.51125)(1316,-3.51357)(1317,-3.52235)(1318,-3.53161)(1319,-3.53928)(1320,-3.54594)(1321,-3.55895)(1322,-3.57378)(1323,-3.57865)(1324,-3.58118)(1325,-3.58779)(1326,-3.5936)(1327,-3.59612)(1328,-3.59997)(1329,-3.60801)(1330,-3.61124)(1331,-3.61463)(1332,-3.61632)(1333,-3.61948)(1334,-3.62216)(1335,-3.62465)(1336,-3.62521)(1337,-3.62665)(1338,-3.6271)(1339,-3.63208)(1340,-3.63339)(1341,-3.64437)(1342,-3.64597)(1343,-3.646)(1344,-3.647)(1345,-3.66024)(1346,-3.66105)(1347,-3.66293)(1348,-3.66945)(1349,-3.6726)(1350,-3.67889)(1351,-3.69781)(1352,-3.70062)(1353,-3.70101)(1354,-3.70893)(1355,-3.71233)(1356,-3.71574)(1357,-3.72094)(1358,-3.74804)(1359,-3.7514)(1360,-3.75142)(1361,-3.75406)(1362,-3.76366)(1363,-3.76973)(1364,-3.77426)(1365,-3.77661)(1366,-3.78876)(1367,-3.80197)(1368,-3.80268)(1369,-3.80474)(1370,-3.8082)(1371,-3.82164)(1372,-3.83079)(1373,-3.83222)(1374,-3.85044)(1375,-3.85192)(1376,-3.86384)(1377,-3.88105)(1378,-3.88285)(1379,-3.88295)(1380,-3.89915)(1381,-3.90476)(1382,-3.91972)(1383,-3.92246)(1384,-3.92941)(1385,-3.9295)(1386,-3.93106)(1387,-3.93366)(1388,-3.95819)(1389,-3.95985)(1390,-3.9602)(1391,-3.96205)(1392,-3.9648)(1393,-3.96495)(1394,-3.96541)(1395,-3.96889)(1396,-3.98265)(1397,-3.99182)(1398,-3.99257)(1399,-3.99968)(1400,-4.00191)(1401,-4.00495)(1402,-4.01444)(1403,-4.01487)(1404,-4.02333)(1405,-4.02999)(1406,-4.03463)(1407,-4.03772)(1408,-4.04148)(1409,-4.04275)(1410,-4.04724)(1411,-4.06002)(1412,-4.06458)(1413,-4.0673)(1414,-4.0674)(1415,-4.0688)(1416,-4.07668)(1417,-4.0772)(1418,-4.07723)(1419,-4.07777)(1420,-4.07874)(1421,-4.08069)(1422,-4.08309)(1423,-4.08477)(1424,-4.08517)(1425,-4.08741)(1426,-4.09233)(1427,-4.09615)(1428,-4.09745)(1429,-4.10099)(1430,-4.1067)(1431,-4.10713)(1432,-4.10871)(1433,-4.11162)(1434,-4.11714)(1435,-4.11779)(1436,-4.12046)(1437,-4.12172)(1438,-4.12541)(1439,-4.13003)(1440,-4.13058)(1441,-4.13721)(1442,-4.14485)(1443,-4.15259)(1444,-4.17276)(1445,-4.17575)(1446,-4.17587)(1447,-4.17985)(1448,-4.1919)(1449,-4.19965)(1450,-4.20151)(1451,-4.21047)(1452,-4.21209)(1453,-4.21346)(1454,-4.21466)(1455,-4.21659)(1456,-4.2229)(1457,-4.22559)(1458,-4.23202)(1459,-4.23217)(1460,-4.23343)(1461,-4.24045)(1462,-4.24329)(1463,-4.25003)(1464,-4.25198)(1465,-4.25619)(1466,-4.25692)(1467,-4.26108)(1468,-4.26791)(1469,-4.26913)(1470,-4.2742)(1471,-4.27696)(1472,-4.29918)(1473,-4.30047)(1474,-4.3085)(1475,-4.31037)(1476,-4.31314)(1477,-4.33575)(1478,-4.33822)(1479,-4.35542)(1480,-4.35859)(1481,-4.35906)(1482,-4.35931)(1483,-4.36168)(1484,-4.36727)(1485,-4.37267)(1486,-4.38528)(1487,-4.39079)(1488,-4.39672)(1489,-4.39946)(1490,-4.40337)(1491,-4.41276)(1492,-4.41432)(1493,-4.41443)(1494,-4.41456)(1495,-4.41765)(1496,-4.4189)(1497,-4.42645)(1498,-4.42802)(1499,-4.42939)(1500,-4.43314)(1501,-4.45799)(1502,-4.45862)(1503,-4.45899)(1504,-4.46931)(1505,-4.47413)(1506,-4.47562)(1507,-4.47671)(1508,-4.49264)(1509,-4.49483)(1510,-4.50025)(1511,-4.50105)(1512,-4.50654)(1513,-4.50685)(1514,-4.51053)(1515,-4.52953)(1516,-4.54119)(1517,-4.54221)(1518,-4.54514)(1519,-4.54664)(1520,-4.54777)(1521,-4.55569)(1522,-4.57255)(1523,-4.57369)(1524,-4.57389)(1525,-4.57829)(1526,-4.58308)(1527,-4.58725)(1528,-4.589)(1529,-4.5891)(1530,-4.59105)(1531,-4.59711)(1532,-4.61081)(1533,-4.61447)(1534,-4.61575)(1535,-4.62039)(1536,-4.62647)(1537,-4.62854)(1538,-4.6382)(1539,-4.64586)(1540,-4.65412)(1541,-4.65895)(1542,-4.66767)(1543,-4.66805)(1544,-4.66997)(1545,-4.66999)(1546,-4.67357)(1547,-4.67835)(1548,-4.6912)(1549,-4.69167)(1550,-4.69644)(1551,-4.69686)(1552,-4.69761)(1553,-4.70184)(1554,-4.70516)(1555,-4.71699)(1556,-4.71735)(1557,-4.71912)(1558,-4.7279)(1559,-4.73103)(1560,-4.73347)(1561,-4.73483)(1562,-4.73756)(1563,-4.73972)(1564,-4.74181)(1565,-4.74464)(1566,-4.75545)(1567,-4.75691)(1568,-4.75977)(1569,-4.76011)(1570,-4.76028)(1571,-4.76128)(1572,-4.76498)(1573,-4.76973)(1574,-4.77579)(1575,-4.77972)(1576,-4.78128)(1577,-4.78667)(1578,-4.78834)(1579,-4.7888)(1580,-4.79948)(1581,-4.80032)(1582,-4.8149)(1583,-4.81838)(1584,-4.83608)(1585,-4.84221)(1586,-4.84572)(1587,-4.8543)(1588,-4.8559)(1589,-4.85935)(1590,-4.86431)(1591,-4.86561)(1592,-4.8686)(1593,-4.87105)(1594,-4.87579)(1595,-4.87665)(1596,-4.88166)(1597,-4.88269)(1598,-4.88955)(1599,-4.89765)(1600,-4.90316)(1601,-4.90601)(1602,-4.91409)(1603,-4.91817)(1604,-4.92071)(1605,-4.92614)(1606,-4.92797)(1607,-4.94296)(1608,-4.94356)(1609,-4.94542)(1610,-4.94824)(1611,-4.94847)(1612,-4.94999)(1613,-4.95134)(1614,-4.9521)(1615,-4.95369)(1616,-4.96943)(1617,-4.97695)(1618,-4.9777)(1619,-4.98716)(1620,-4.98844)(1621,-4.98979)(1622,-4.99469)(1623,-4.99528)(1624,-4.99926)(1625,-4.99986)(1626,-5.00032)(1627,-5.00132)(1628,-5.00481)(1629,-5.00757)(1630,-5.01107)(1631,-5.01127)(1632,-5.02002)(1633,-5.02293)(1634,-5.03454)(1635,-5.03483)(1636,-5.03729)(1637,-5.03735)(1638,-5.03935)(1639,-5.04641)(1640,-5.04741)(1641,-5.04752)(1642,-5.05081)(1643,-5.05426)(1644,-5.05855)(1645,-5.06727)(1646,-5.0694)(1647,-5.07259)(1648,-5.0729)(1649,-5.07399)(1650,-5.08492)(1651,-5.08707)(1652,-5.08867)(1653,-5.0911)(1654,-5.09149)(1655,-5.09839)(1656,-5.0989)(1657,-5.10062)(1658,-5.10367)(1659,-5.10867)(1660,-5.11157)(1661,-5.11311)(1662,-5.11354)(1663,-5.11755)(1664,-5.12006)(1665,-5.12063)(1666,-5.1224)(1667,-5.12306)(1668,-5.12451)(1669,-5.14111)(1670,-5.14531)(1671,-5.1478)(1672,-5.15493)(1673,-5.16012)(1674,-5.16112)(1675,-5.16232)(1676,-5.16594)(1677,-5.16735)(1678,-5.17478)(1679,-5.17529)(1680,-5.18024)(1681,-5.18225)(1682,-5.18344)(1683,-5.18977)(1684,-5.193)(1685,-5.19463)(1686,-5.20225)(1687,-5.20283)(1688,-5.20424)(1689,-5.20757)(1690,-5.21764)(1691,-5.22276)(1692,-5.22347)(1693,-5.22837)(1694,-5.22871)(1695,-5.24048)(1696,-5.24298)(1697,-5.24411)(1698,-5.24804)(1699,-5.24932)(1700,-5.24984)(1701,-5.25164)(1702,-5.25383)(1703,-5.25921)(1704,-5.25938)(1705,-5.26098)(1706,-5.2649)(1707,-5.2676)(1708,-5.27268)(1709,-5.27529)(1710,-5.2852)(1711,-5.28798)(1712,-5.29582)(1713,-5.30015)(1714,-5.30724)(1715,-5.3076)(1716,-5.31239)(1717,-5.32438)(1718,-5.32721)(1719,-5.32778)(1720,-5.3281)(1721,-5.34279)(1722,-5.3572)(1723,-5.35808)(1724,-5.36065)(1725,-5.36761)(1726,-5.36918)(1727,-5.37104)(1728,-5.37304)(1729,-5.38314)(1730,-5.38588)(1731,-5.39259)(1732,-5.39437)(1733,-5.39745)(1734,-5.39805)(1735,-5.39874)(1736,-5.40009)(1737,-5.40852)(1738,-5.41089)(1739,-5.41115)(1740,-5.41907)(1741,-5.42737)(1742,-5.42936)(1743,-5.43165)(1744,-5.43963)(1745,-5.44101)(1746,-5.4434)(1747,-5.45646)(1748,-5.45876)(1749,-5.46247)(1750,-5.4673)(1751,-5.46765)(1752,-5.47079)(1753,-5.47749)(1754,-5.4776)(1755,-5.48083)(1756,-5.48531)(1757,-5.49406)(1758,-5.49698)(1759,-5.49887)(1760,-5.50155)(1761,-5.50358)(1762,-5.51354)(1763,-5.51512)(1764,-5.51917)(1765,-5.52212)(1766,-5.52298)(1767,-5.52575)(1768,-5.5275)(1769,-5.53024)(1770,-5.53696)(1771,-5.54227)(1772,-5.54371)(1773,-5.54737)(1774,-5.5506)(1775,-5.55091)(1776,-5.55476)(1777,-5.55751)(1778,-5.55828)(1779,-5.56138)(1780,-5.56424)(1781,-5.57715)(1782,-5.58545)(1783,-5.58665)(1784,-5.59157)(1785,-5.59345)(1786,-5.59466)(1787,-5.59505)(1788,-5.59511)(1789,-5.59692)(1790,-5.5972)(1791,-5.61301)(1792,-5.61653)(1793,-5.62122)(1794,-5.62524)(1795,-5.62559)(1796,-5.62564)(1797,-5.63345)(1798,-5.63714)(1799,-5.64234)(1800,-5.64301)(1801,-5.65272)(1802,-5.65286)(1803,-5.66102)(1804,-5.66241)(1805,-5.66408)(1806,-5.66596)(1807,-5.67279)(1808,-5.67496)(1809,-5.67857)(1810,-5.68329)(1811,-5.69209)(1812,-5.69341)(1813,-5.69367)(1814,-5.69476)(1815,-5.69602)(1816,-5.69741)(1817,-5.70414)(1818,-5.71031)(1819,-5.71303)(1820,-5.71326)(1821,-5.71657)(1822,-5.72389)(1823,-5.72446)(1824,-5.72569)(1825,-5.72609)(1826,-5.74324)(1827,-5.74501)(1828,-5.74776)(1829,-5.74828)(1830,-5.75377)(1831,-5.75603)(1832,-5.77078)(1833,-5.77693)(1834,-5.77956)(1835,-5.78081)(1836,-5.78847)(1837,-5.7925)(1838,-5.79308)(1839,-5.79711)(1840,-5.80318)(1841,-5.80457)(1842,-5.80689)(1843,-5.80867)(1844,-5.81006)(1845,-5.81713)(1846,-5.81838)(1847,-5.8233)(1848,-5.82539)(1849,-5.8309)(1850,-5.83463)(1851,-5.83534)(1852,-5.83652)(1853,-5.84298)(1854,-5.84428)(1855,-5.84958)(1856,-5.85241)(1857,-5.8539)(1858,-5.85641)(1859,-5.85924)(1860,-5.86211)(1861,-5.87937)(1862,-5.88801)(1863,-5.89195)(1864,-5.89283)(1865,-5.89841)(1866,-5.89869)(1867,-5.89882)(1868,-5.89927)(1869,-5.89935)(1870,-5.89962)(1871,-5.90713)(1872,-5.91216)(1873,-5.92563)(1874,-5.92714)(1875,-5.92854)(1876,-5.9304)(1877,-5.93287)(1878,-5.93879)(1879,-5.94045)(1880,-5.94184)(1881,-5.94239)(1882,-5.94498)(1883,-5.94613)(1884,-5.94851)(1885,-5.9488)(1886,-5.95349)(1887,-5.95863)(1888,-5.96298)(1889,-5.9632)(1890,-5.96392)(1891,-5.96411)(1892,-5.96795)(1893,-5.96867)(1894,-5.9701)(1895,-5.9723)(1896,-5.97725)(1897,-5.98754)(1898,-5.99226)(1899,-5.99437)(1900,-6.00002)(1901,-6.00341)(1902,-6.01167)(1903,-6.01187)(1904,-6.01383)(1905,-6.02359)(1906,-6.02656)(1907,-6.0327)(1908,-6.0382)(1909,-6.04289)(1910,-6.04531)(1911,-6.04601)(1912,-6.05773)(1913,-6.05804)(1914,-6.05832)(1915,-6.05955)(1916,-6.0637)(1917,-6.06508)(1918,-6.06604)(1919,-6.06885)(1920,-6.06922)(1921,-6.07351)(1922,-6.07443)(1923,-6.07579)(1924,-6.07596)(1925,-6.07717)(1926,-6.08038)(1927,-6.08072)(1928,-6.0872)(1929,-6.092)(1930,-6.09212)(1931,-6.09495)(1932,-6.09764)(1933,-6.10354)(1934,-6.10391)(1935,-6.10854)(1936,-6.10974)(1937,-6.1108)(1938,-6.11333)(1939,-6.11955)(1940,-6.12341)(1941,-6.12375)(1942,-6.12718)(1943,-6.12857)(1944,-6.1291)(1945,-6.1305)(1946,-6.13057)(1947,-6.13316)(1948,-6.13518)(1949,-6.13781)(1950,-6.13885)(1951,-6.14617)(1952,-6.14874)(1953,-6.15265)(1954,-6.15845)(1955,-6.15869)(1956,-6.15905)(1957,-6.16203)(1958,-6.164)(1959,-6.16527)(1960,-6.16566)(1961,-6.1682)(1962,-6.17124)(1963,-6.17705)(1964,-6.17818)(1965,-6.17864)(1966,-6.18291)(1967,-6.18373)(1968,-6.18467)(1969,-6.18756)(1970,-6.19371)(1971,-6.19451)(1972,-6.19491)(1973,-6.19665)(1974,-6.19772)(1975,-6.19837)(1976,-6.213)(1977,-6.2179)(1978,-6.21813)(1979,-6.22033)(1980,-6.22304)(1981,-6.22479)(1982,-6.22976)(1983,-6.22996)(1984,-6.23319)(1985,-6.23942)(1986,-6.25126)(1987,-6.25129)(1988,-6.25553)(1989,-6.26081)(1990,-6.26095)(1991,-6.26596)(1992,-6.2714)(1993,-6.27214)(1994,-6.27642)(1995,-6.27746)(1996,-6.28137)(1997,-6.28198)(1998,-6.28674)(1999,-6.28677)(2000,-6.29679) 
};

\addplot [
color=mycolor2,
solid,
line width=1.0pt
]
coordinates{
 (0,0.0)(2000,0.0) 
};

\addplot [
color=mycolor3,
dashed,
line width=1.0pt
]
coordinates{
 (0,-0.474609)(2000,-0.474609) 
};
\end{axis}
\end{tikzpicture}

%% file: habigt2013.bbl
\begin{thebibliography}{10}

\bibitem{Smolic2011}
A.~Smolic, P.~Kauff, S.~Knorr, A.~Hornung, M.~Kunter, M.~M\"{u}ller, and
  M.~Lang,
\newblock ``Three-dimensional video postproduction and processing,''
\newblock {\em Proc. IEEE}, vol. 99, no. 4, pp. 607--625, Apr. 2011.

\bibitem{Held2008}
R.~Held and M.~Banks,
\newblock ``{Misperceptions in Stereoscopic Displays: A Vision Science
  Perspective},''
\newblock in {\em Proc. 5th Symp. Applied Perception Graphics and
  Visualization}, Los Angeles, CA, 2008, pp. 23--32.

\bibitem{Chen1993}
S.E. Chen and L.~Williams,
\newblock ``{View interpolation for image synthesis},''
\newblock in {\em Proc. 20th Annu. Conf. and Exhibition Computer Graphics and
  Interactive Techniques}, Anaheim, CA, 1993, pp. 279--288.

\bibitem{Fehn2004}
C.~Fehn,
\newblock ``Depth-image-based rendering {(DIBR)}, compression, and transmission
  for a new approach on {3D-TV},''
\newblock in {\em Proc. {SPIE}}, San Jose, CA, 2004, vol. 5291, pp. 93--104.

\bibitem{Zhang2005}
L.~Zhang and W.J. Tam,
\newblock ``{Stereoscopic Image Generation Based on Depth Images for 3D TV},''
\newblock {\em IEEE Trans. Broadcast.}, vol. 51, no. 2, pp. 191--199, June
  2005.

\bibitem{Daribo2010}
I.~Daribo and H.~Saito,
\newblock ``Bilateral depth-discontinuity filter for novel view synthesis,''
\newblock in {\em Proc. IEEE Int. Workshop Multimedia Signal Processing},
  Saint-Malo, France, 2010, pp. 145--149.

\bibitem{Criminisi2004}
A.~Criminisi, P.~Perez, and K.~Toyama,
\newblock ``Region filling and object removal by exemplar-based image
  inpainting,''
\newblock {\em IEEE Trans. Image Process.}, vol. 13, no. 9, pp. 1200--1212,
  Sept. 2004.

\bibitem{Daribo2011}
I.~Daribo and H.~Saito,
\newblock ``{A Novel Inpainting-Based Layered Depth Video for 3DTV},''
\newblock {\em IEEE Trans. Broadcast.}, vol. 57, no. 2, pp. 533--541, June
  2011.

\bibitem{Oh2009}
K.~Oh, S.~Yea, and Y.~Ho,
\newblock ``Hole filling method using depth based in-painting for view
  synthesis in free viewpoint television and {3-D} video,''
\newblock in {\em Proc. 27th Picture Coding Symp.}, Chicago, IL, 2009, pp.
  1--4.

\bibitem{Gautier2011}
J.~Gautier, O.~Le~Meur, and C.~Guillemot,
\newblock ``{Depth-based image completion for view synthesis},''
\newblock in {\em Proc. 3DTV Conf. The True Vision Capture Transmission and
  Display of 3D Video}, Antalya, Turkey, 2011, pp. 1--4.

\bibitem{Komodakis2007}
N.~Komodakis and G.~Tziritas,
\newblock ``Image completion using efficient belief propagation via priority
  scheduling and dynamic pruning,''
\newblock {\em IEEE Trans. Image Process.}, vol. 16, no. 11, pp. 2649--2661,
  Nov. 2007.

\bibitem{Zitnick2004}
C.L. Zitnick, S.B. Kang, and M.~Uyttendaele,
\newblock ``High-quality video view interpolation using a layered
  representation,''
\newblock {\em ACM Trans. Graphics}, vol. 23, no. 3, pp. 600--608, Aug. 2004.

\bibitem{VSRS2008}
M.~Tanimoto, T.~Fujii, and K.~Suzuki,
\newblock ``View synthesis algorithm in {V}iew {S}ynthesis {R}eference
  {S}oftware 2.0 ({VSRS2.0}),''
\newblock Tech. {R}ep., ISO/IEC JTC1/SC29/WG11 M16090, Lausanne, Switzerland,
  Feb. 2008.

\end{thebibliography}
